\documentclass[letterpaper]{article} 
\usepackage{aaai2027}  
\usepackage[hyphens]{url}  
\usepackage{graphicx} 
\usepackage{natbib}  
\usepackage{caption} 
\usepackage{algorithm}
\usepackage{algorithmic}
\usepackage{amsmath}
\usepackage{newfloat}
\usepackage{listings}
\DeclareCaptionStyle{ruled}{labelfont=normalfont,labelsep=colon,strut=off} 
\floatstyle{ruled}
\newfloat{listing}{tb}{lst}{}
\floatname{listing}{Listing}

\usepackage{booktabs}
\usepackage{subcaption}
\usepackage{graphicx}
\usepackage{amssymb}
\usepackage[table]{xcolor}
\newcommand{\cmark}{\checkmark}
\usepackage[most]{tcolorbox}
\usepackage{enumitem}
\usepackage{tabularx}
\usepackage{array}
\nocopyright

\title{StatePlay: State-Aware Game World Models for Mechanics-Consistent Generation}
\author{
    Anonymous submission
}
\affiliations{
    Anonymous institution
}

\title{StatePlay: State-Aware Game World Models for Mechanics-Consistent Generation}
\author{
    Zijun Lin\textsuperscript{\rm 1,\rm 2,4,\ensuremath{\dagger}},
    Zeqing Wang\textsuperscript{\rm 1,\rm 3,\ensuremath{\dagger}},
    Cheston Tan\textsuperscript{\rm 4},
    Bihan Wen\textsuperscript{\rm 2},
    Yeying Jin\textsuperscript{\rm 1,\rm 3,\ensuremath{\ddagger},}\corresponding
}
\affiliations {
    \textsuperscript{\rm 1}Tencent
    \textsuperscript{\rm 2}Nanyang Technological University\\
    \textsuperscript{\rm 3}National University of Singapore
    \textsuperscript{\rm 4}Centre for Frontier AI Research, A*STAR\\
}

\begin{document}

\maketitle

\begingroup
\renewcommand{\thefootnote}{\ensuremath{\dagger}}
\footnotetext{This work was completed during research internships at Tencent under the supervision of Yeying Jin.}
\endgroup

\begingroup
\renewcommand{\thefootnote}{\ensuremath{\ddagger}}
\footnotetext{Project Lead.}
\endgroup

\begin{abstract}
Recent game world models can generate visually realistic and interactive environments conditioned on player actions. However, games are not defined by pixels alone; they are governed by explicit mechanics, namely state-dependent rules that control health reduction, skill activation, and game termination. These mechanics depend on precise internal states, such as health points, skill meters, and timers, which are tightly coupled with visual observations and determine how gameplay evolves. Without modeling these state dynamics, existing game world models may generate visually plausible rollouts but violate the underlying game rules. In this paper, we propose \textbf{StatePlay}, a novel state-aware game world model that jointly predicts visual content and game states to promote mechanics-consistent generation. StatePlay adopts a mixture-of-transformers (MoT)-style architecture that preserves specialized visual and state representations while enabling cross-modal interaction, allowing predicted states to guide frame generation. Each branch is further optimized with a distinct objective suited to its modality. Experiments show that StatePlay achieves an average normalized L1 distance below 0.06 for state prediction. Furthermore, compared with models without explicit state modeling, our method improves mechanics fidelity in generated game rollouts by 18.6\%. Overall, our work highlights the importance of state-aware game world modeling and advances beyond pixel-level realism toward complete and mechanically faithful game generation. Project Page: \textcolor{cyan}{https://jimntu.github.io/stateplay\_page/}

\end{abstract}


\section{Introduction}


\begin{figure}[t]
    \centering
    \includegraphics[width=0.99\columnwidth]{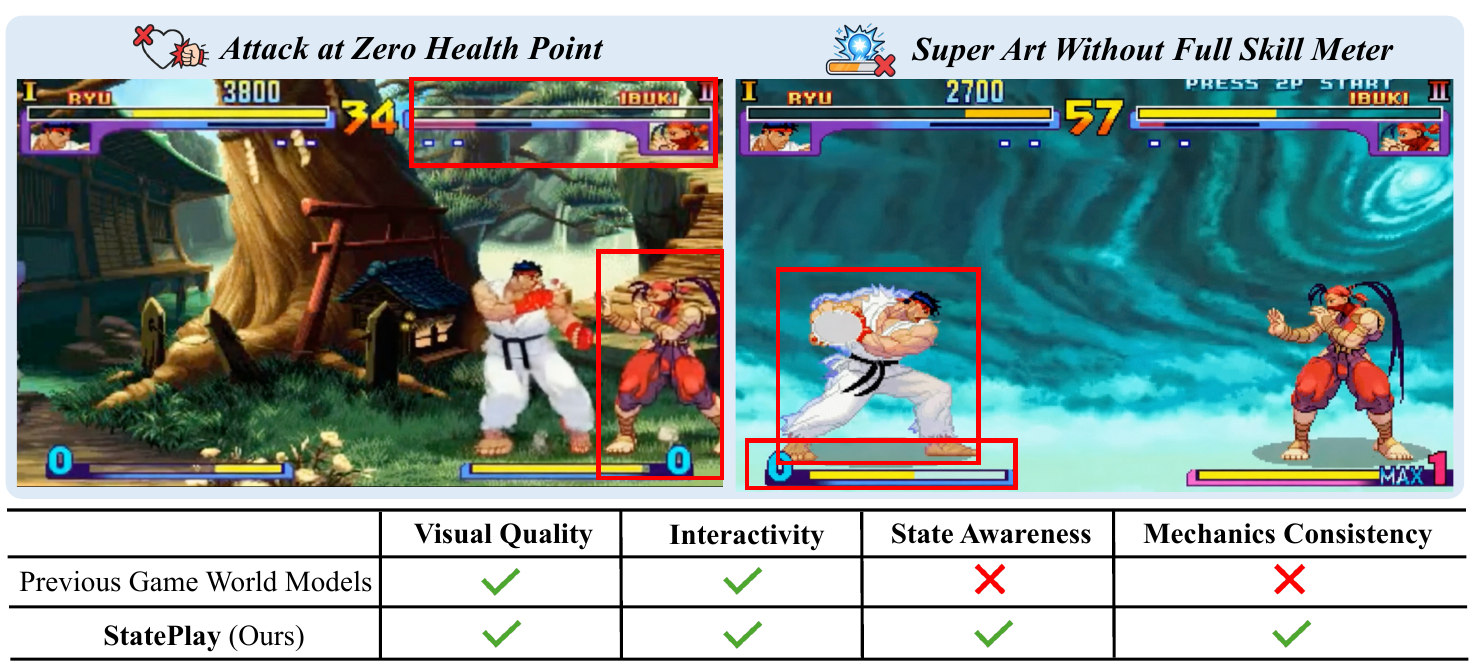}
    \caption{
    Existing game world models produce mechanically inconsistent gameplay, such as attacking at zero health or triggering a super art without a full skill meter (\textcolor{red}{red} boxes, top). In contrast, \textbf{StatePlay} jointly achieves state awareness and mechanics consistency (bottom table).
    }
    \label{fig:compare}
\end{figure}

World models have attracted increasing attention for simulating future environment dynamics conditioned on agent or user actions \cite{bruce2024genie,mao2026yume1}. Unlike traditional video generation models \cite{seedance2026seedance,team2025kling}, they emphasize action-conditioned prediction, where future observations are jointly determined by historical context and interactive inputs. This capability makes them well suited to closed-loop applications such as autonomous driving \cite{wang2024drivedreamer}, embodied AI \cite{ye2026world}, and games \cite{yu2025gamefactory}. In particular, games provide a promising testbed because the environment must respond coherently to player actions over time. Recent game world models can generate visually realistic and controllable gameplay rollouts while supporting interactions with dynamic environments \cite{tong2026scope}, non-player characters \cite{wang2026reactivegwm}, and multiple players \cite{wu2026multiworld}, enabling more immersive, creative, and engaging gameplay experiences.

\begin{figure*}[t!]
\centering
    \includegraphics[width=0.99\textwidth]{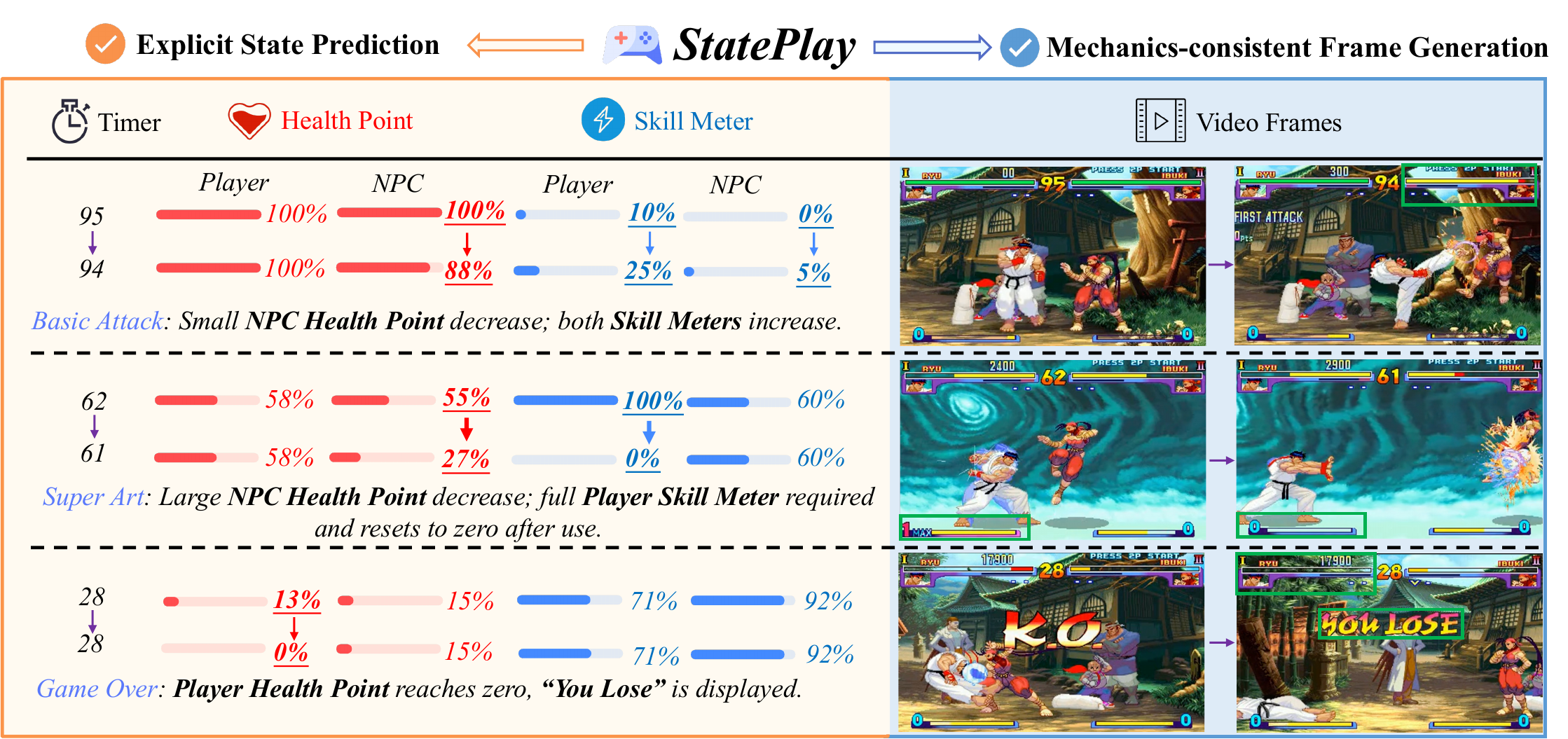}
    \caption{
    Overview of \textbf{StatePlay}. By explicitly modeling game states, including timers, health points, and skill meters, StatePlay guides frame generation to remain consistent with the underlying game mechanics (highlighted in \textcolor{green!50!black}{green} bounding box).
    }
    \label{fig:teaser}
\end{figure*}

However, simulating a playable game requires more than generating realistic frames that follow user intent. Games are governed by underlying mechanics, which are often determined by internal state variables such as health points, skill meters, and timers. These states regulate valid actions, visual transitions, and game progression. Taking fighting games as an example, a player can perform certain special attacks only when the skill meter is full, and the game should end once either player’s health points reach zero. Therefore, valid and consistent mechanics are fundamental to games, directly determining whether a generated environment is truly playable or merely a visual demo.

Despite the importance of internal states, existing game world models formulate gameplay generation as a pixel-space prediction task and overlook the causal role of state dynamics in shaping future observations. Consequently, current models may generate visually plausible rollouts but fail to follow the game rules. As shown in Fig. \ref{fig:compare}, players may continue fighting after the game should have ended, or perform super arts without full skill meters. Such failures lead to mechanically inconsistent and incomplete gameplay experiences.

To address this gap, we introduce \textbf{StatePlay}, the first state-aware game world model that jointly predicts game states and visual content for mechanics-consistent gameplay generation (See Fig. \ref{fig:teaser}). Our key observation is that player actions are often directly reflected in pixel-level changes, whereas internal state dynamics cannot be reliably learned from visual observations alone. Therefore, existing game world model datasets that contain only frames and actions are insufficient for learning state transitions and game rules.

Based on this observation, we develop StatePlay from both the data and model perspectives. First, due to the limited availability of open-source datasets with explicit state annotations, we construct a synchronized state–frame–action dataset from \textit{Street Fighter 3} (SF3) \cite{capcom1998streetfighteralpha3}. The dataset records game frames and player actions together with internal states, including health points, skill meters, and timers, enabling us to systematically investigate the effect of explicit state modeling on game world models. Second, inspired by the World Action Model (WAM) \cite{ye2026world, kim2026cosmos}, which jointly models visual frames and actions, we design StatePlay with a mixture-of-transformers (MoT)-style backbone that simultaneously predicts game states and generates visual content. Specifically, StatePlay couples the two modalities while optimizing each branch with a tailored objective to produce rollouts that are both visually plausible and consistent with game mechanics.

Experiments highlight the importance of generating visual content with explicit state prediction. StatePlay accurately models internal game states, achieving an average normalized L1 distance below 0.06 across key state variables. Beyond state prediction itself, coupling states with visual generation leads to a 18.6\% improvement in mechanics fidelity over the best-performing baseline. Compared with stateless game world models that learn game mechanics implicitly through pixel-level supervision, StatePlay better preserves state-dependent events, such as skill activation, health reduction, and game termination, while maintaining visual quality and action controllability. These results indicate that explicit state modeling is crucial for advancing game world models beyond visual realism toward more playable and mechanically consistent simulation.

Overall, our contributions are summarized as follows:
\begin{itemize}
\item We propose \textbf{StatePlay}, the first state-aware game world model that jointly predicts game states and visual content, addressing the lack of state modeling in prior work.
\item We introduce a novel mixture-of-transformers (MoT)-style architecture with distinct objectives for the state and visual branches. By explicitly coupling state modeling with visual generation, it enables accurate state prediction and mechanically consistent frame generation.
\item We conduct extensive experiments demonstrating that StatePlay improves mechanics fidelity by 18.6\% over stateless game world models, while maintaining accurate state prediction, action controllability, and visual quality.
\end{itemize}

\section{Related Work}
\subsection{Game World Model}
Game world models have the potential to transform game development by autonomously generating interactive content conditioned on player actions. Pioneering work such as Genie \cite{bruce2024genie} demonstrates that video generative models can produce playable environments by learning action-controllable dynamics from unlabeled videos. Studies such as ReactiveGWM \cite{wang2026reactivegwm}, Incantation \cite{zhu2026incantation}, and MultiWorld \cite{wu2026multiworld} further extend this direction toward richer forms of interactivity, from free exploration to interactions with environments, NPCs, and other players. In parallel, recent works including Lyra 2.0 \cite{shen2026lyra}, HY-World 1.5 \cite{sun2025worldplay}, and Matrix-Game 3.0 \cite{wang2026matrix} focus on improving real-time interaction, long-horizon consistency, and high-resolution generation. 

Despite rapid progress in game world models, state prediction, which is equally important for building playable environments, remains largely overlooked. In this work, we identify this missing component in recent game world model research and enable game world models to simultaneously predict game states and generate video frames that follow the underlying game rules.

\subsection{State-Aware World Modeling}
The use of state information has been widely studied in embodied and decision-making domains, where the definition of state depends on the task setting. In embodied AI, state often refers to proprioceptive information, such as robot poses and joint configurations, which helps policies generate physically feasible actions \cite{black2025pi_,kimopenvla}. In autonomous driving, state variables such as ego speed, acceleration, and surrounding object motion are commonly modeled to support safer future prediction and planning \cite{zheng2024occworld,hu2022model}. These examples suggest that explicit state modeling is crucial for systems that require both visual understanding and rule-consistent interaction.

Games also involve task-dependent states, such as ammunition or bandage counts in first-person shooter (FPS) games, nitro or speed in racing games, and health points or skill meters in fighting games. However, this perspective remains underexplored in game world models because game state information cannot always be directly extracted from the environment or reliably inferred from generated content. The most closely related recent works, including WildWorld \cite{li2026wildworld} and From Pixels to States \cite{li2026pixelsstatesrethinkinginteractive}, introduce action-conditioned video datasets with explicit state annotations, but it remains unclear how state prediction can improve the mechanics fidelity of the frame generation, or how such state information should be incorporated into game world models. In this work, we propose StatePlay, a state-aware game world model, to fill this gap and systematically investigate how visual generation and state prediction can complement and reinforce each other.


\section{StatePlay}
\subsection{State-Aware Dataset Construction}
We specifically construct our dataset using \textit{Street Fighter 3} (SF3) \cite{capcom1998streetfighteralpha3}, as few publicly available game datasets provide explicit state annotations. This game offers a suitable starting point for studying the relationship between state prediction and mechanics-consistent visual generation, as it contains intuitive state variables and clearly defined game rules. These states are directly tied to game progression: different attacks cause different amounts of health-point reduction, super arts can be activated only when the skill meter is full, and the game ends when either character's health reaches zero. The state-aware dataset is constructed in three stages: gameplay recording, mechanics distribution balancing, and NPC strategy annotation.

\subsubsection{Gameplay Recording}
Following a similar data acquisition pipeline to ReactiveGWM \cite{wang2026reactivegwm}, the stable-retro framework \cite{poliquin2026stableretro} is employed to programmatically collect gameplay episodes. It provides a Gymnasium-compatible interface for SF3, allowing us to control the game through Python, record visual observations and actions. Players are controlled by agents that sample actions from an 11-dimensional action space, including 4 movement actions, 6 normal attacks, and 1 super art. Additionally, we extract five state variables from emulator memory: the game timer, the health points and the skill meters of both the player and the NPC. Each episode runs until a round-ending knock-out, and is then segmented into 5-second clips at 20 FPS. 


\subsubsection{Mechanics Distribution Balancing}
\label{subsubsec: mechanics}
After collecting the gameplay clips with synchronized player actions and internal states, we balance the dataset across five mechanics-related categories: \textit{result win}, \textit{result lose}, \textit{macro success}, \textit{macro fail}, and \textit{normal}. Labels are assigned using both Gemini-3.1-Pro \cite{google2026gemini31pro} and the recorded game states. 

Specifically, a clip is labeled as \textit{result win} or \textit{result lose} when Gemini detects the corresponding ``You Win'' or ``You Lose'' frame, and one character reaches zero health. A clip is labeled as \textit{macro success} when Gemini detects a successfully executed super art, the recorded action contains the corresponding command, and the skill meter exceeds the required threshold. In contrast, it is labeled as \textit{macro fail} when the player issues the command with insufficient skill meter, causing a regular attack instead. The remaining clips without these state-dependent events are labeled as \textit{normal}.

Finally, we construct a balanced training set with 10,000 clips. State-critical cases account for 40\% of the dataset, including \textit{result win}, \textit{result lose}, \textit{macro success}, and \textit{macro fail}, each contributing 10\%. The remaining 60\% consists of \textit{normal} cases. This balancing strategy ensures that the dataset contains sufficient clips where state information is essential for generating mechanics-consistent game frames, while still preserving common gameplay dynamics. Please refer to Appendix \ref{subsec:mechanics_distribution} for more details.

\begin{figure*}[t!]
\centering
    \includegraphics[width=0.99\textwidth]{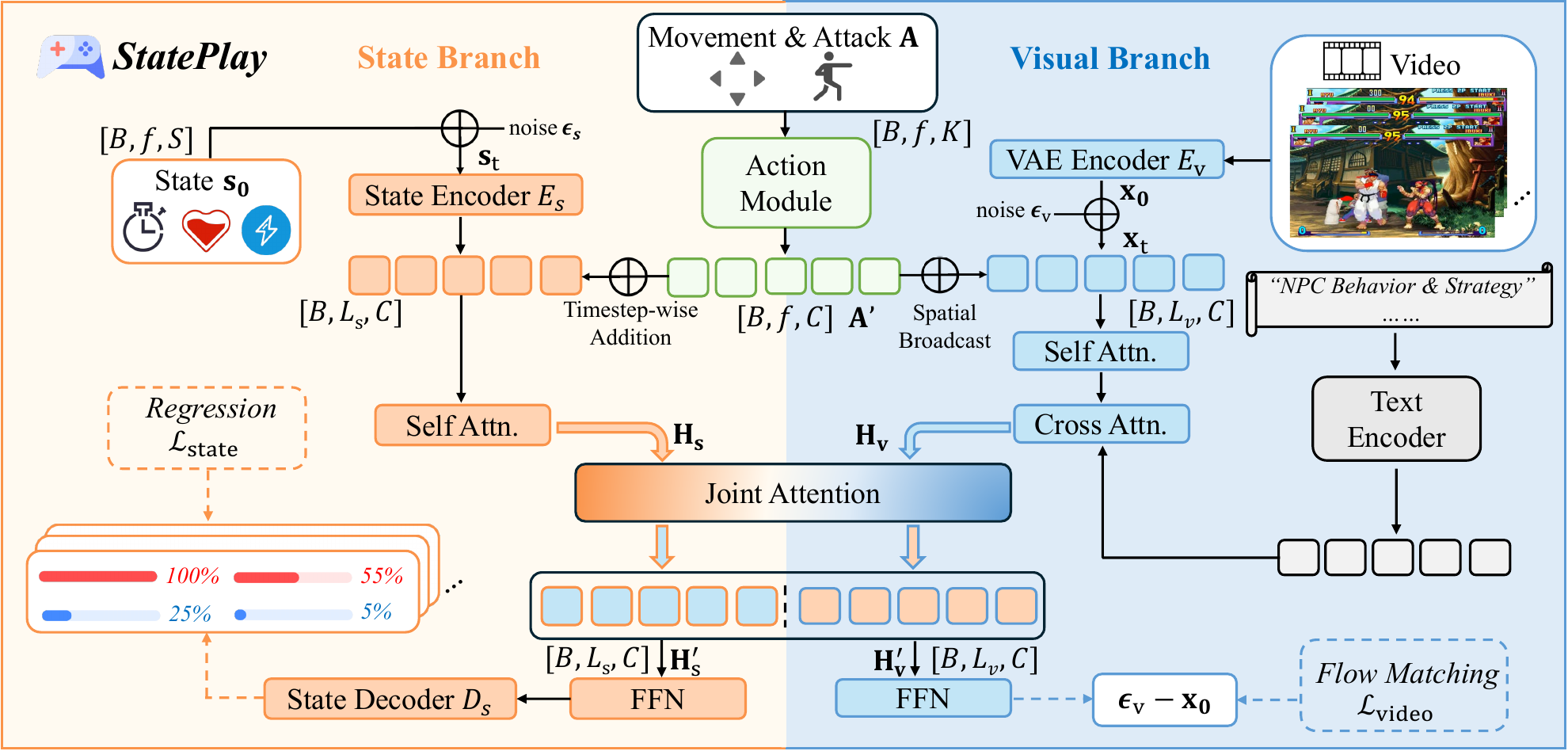}
    \caption{
\textbf{StatePlay} block architecture during training. StatePlay adopts a Mixture-of-Transformers (MoT)-style architecture with separate state and visual branches. A shared joint-attention module enables information exchange between the two branches. The state branch is optimized with a regression loss, while the visual branch is trained using a flow-matching objective. White-filled boxes denote inputs and supervision targets, while colored boxes represent model components. The block is repeated $N$ times.}
    \label{fig:model_arch}
\end{figure*}

\subsubsection{NPC Strategy Annotation}
To enable the model to steer the NPC toward different strategies against the player, we use Gemini-3.1-Pro to annotate NPC behavior with natural-language descriptions. Following \cite{wang2026reactivegwm}, we design the prompt to describe NPC behavior along three strategy categories: \textit{Offense}, where the NPC closes the distance and attacks the player proactively; \textit{Control}, where the NPC maintains distance and uses projectiles; and \textit{Defense}, where the NPC reacts passively with crouching guards. Please refer to the Appendix \ref{subsec:NPC_anno} for the detailed NPC strategy prompt.

Altogether, we construct a state-aware dataset with 10,000 training clips including video frames, player actions, state information, and an NPC textual description, with a balanced distribution across different mechanics-related scenarios.

\subsection{Model Architecture}
\subsubsection{State and Visual Branches}
To jointly predict internal game states and generate mechanics-consistent visual frames, we introduce a lightweight state branch (0.76B) alongside the visual branch (5B) commonly adopted by existing game world models. This simple yet effective design shown in Fig. \ref{fig:model_arch} can be readily integrated into different game world models to endow them with state modeling capability.

Inspired by the action-conditioning design in world action models, we construct the state branch by perturbing the clean state sequence $\mathbf{s}_0$ with Gaussian noise $\boldsymbol{\epsilon}_s$ according to a sampled timestep $t$:

\begin{equation}
\mathbf{s}_t
=
(1-t)\mathbf{s}_0
+
t\boldsymbol{\epsilon}_s,
\qquad
\boldsymbol{\epsilon}_s
\sim
\mathcal{N}(\mathbf{0},\mathbf{I}),
\end{equation}
where $\mathbf{s}_0 \in \mathbb{R}^{B \times f \times S}$ with $B$ denotes the batch size, $f$ denotes the number of latent frames, and $S$ denotes the number of state variables, which may vary across different games. 

During inference, the state of the first frame is provided as a condition, while the states of the remaining frames are initialized from noise and predicted by the model. The perturbed state sequence is then processed by the state encoder $E_s$ and state self-attention module to obtain the state representation $\mathbf{H}_s\in \mathbb{R}^{B \times L_s \times C}$, which captures temporal state dynamics and mechanics-related information with $L_S$ state token length containing $C$ dimension each.

Similarly, the video branch perturbs the clean latent representation $\mathbf{x}_0$ produced by the VAE encoder using independently sampled Gaussian noise $\boldsymbol{\epsilon}_v$:
 
\begin{equation}
\mathbf{x}_t
=
(1-t)\mathbf{x}_0
+
t\boldsymbol{\epsilon}_v,
\qquad
\boldsymbol{\epsilon}_v
\sim
\mathcal{N}(\mathbf{0},\mathbf{I}),
\end{equation}
where $\mathbf{x}_0 \in \mathbb{R}^{B \times L_v \times C}$ denotes the visual latent representation, consisting of $L_v$ visual tokens, each with feature dimension $C$, matching the dimension of the state tokens. The perturbed visual latent $\mathbf{x}_t$ is then processed by the visual self-attention module, followed by cross-attention with the text embedding, as illustrated in Fig.~\ref{fig:model_arch}, producing the visual representation $\mathbf{H}_v \in \mathbb{R}^{B \times L_v \times C}$ that captures both visual dynamics and semantic information from the text prompt. The state representation $\mathbf{H}_s$ and visual representation $\mathbf{H}_v$ are processed separately by their respective expert branches, as they correspond to intrinsically different modalities. The effectiveness of this modality-specific design is further analyzed in Sec. \ref{subsubsec:couple}.

In addition, the action sequence $\mathbf{A} \in \mathbb{R}^{B \times f \times K}$, where $K$ denotes the action dimension, is projected by an action encoder into an action embedding $\mathbf{A}' \in \mathbb{R}^{B \times f \times C}$. The action embedding is then injected into both the visual and state branches immediately after token embedding, enabling player actions to jointly modulate visual generation and state prediction throughout the network.

\subsubsection{Mixture of Transformers}
After obtaining the state representation $\mathbf{H}_s$ and visual representation $\mathbf{H}_v$ from their respective expert branches, a joint attention module is applied for cross-modal information exchange:
\begin{equation}
\begin{aligned}
\mathbf{H}_v'
&=
\mathbf{H}_v+
\operatorname{MHA}
(\mathbf{H}_v,\mathbf{H}_s,\mathbf{H}_s),\\
\mathbf{H}_s'
&=
\mathbf{H}_s+
\operatorname{MHA}
(\mathbf{H}_s,\mathbf{H}_v,\mathbf{H}_v).
\end{aligned}
\end{equation}
Specifically, the visual tokens serve as queries and attend to the keys and values derived from the state tokens, allowing visual generation to incorporate mechanics-related state information. Conversely, the state tokens serve as queries and attend to the keys and values derived from the visual tokens, enabling state prediction to leverage the corresponding visual dynamics. This bidirectional interaction enables effective information exchange while preserving modality-specific representations in each branch.
\subsubsection{Loss Design}

StatePlay is jointly optimized for visual generation and state prediction. For the visual branch, we follow the standard flow matching objective adopted by Wan-based game world models to learn the velocity field for denoising the perturbed visual latent:

\begin{equation}
\mathcal{L}_{\mathrm{video}}
=
\mathbb{E}_{\mathbf{x}_0,\boldsymbol{\epsilon}_v,t}
\left[
\left\|
v_{\theta}(\mathbf{x}_t,t)
-
\left(
\boldsymbol{\epsilon}_v-\mathbf{x}_0
\right)
\right\|_2^2
\right],
\end{equation}
where $v_{\theta}$ denotes the velocity field predicted by the visual branch, $\mathbf{x}_t$ is the perturbed visual latent at timestep $t$, and $\boldsymbol{\epsilon}_v-\mathbf{x}_0$ is the corresponding ground-truth velocity under the linear flow path.

For the state branch, rather than applying the same flow-matching objective used in world action models, we directly supervise the predicted states with a Smooth L1 regression loss. Compared with naively adapting flow matching to state prediction, regression is better suited to the low-dimensional, mechanics-constrained nature of game states:

\begin{equation}
\mathcal{L}_{\mathrm{state}}
=
\operatorname{SmoothL1}
\left(
D_s(\mathbf{H}_s'),
\mathbf{s}_0
\right),
\end{equation}
where $D_s$ denotes the state decoder, and the loss is computed between the predicted state sequence $D_s(\mathbf{H}_s')$ and the corresponding clean state sequence $\mathbf{s}_0$.

Finally, the two objectives are jointly optimized using a weighted summation:

\begin{equation}
\mathcal{L}_{\mathrm{train}}
=
\lambda_{\mathrm{state}}\mathcal{L}_{\mathrm{state}}
+
\lambda_{\mathrm{video}}\mathcal{L}_{\mathrm{video}},
\end{equation}
where $\lambda_{\mathrm{state}}$ and $\lambda_{\mathrm{video}}$ denote the loss weights for the state and visual branches, respectively. A detailed ablation study of this design choice is provided in Sec. \ref{subsubsec:state_loss}.
\section{Experiments}
\subsection{Evaluation Metrics}

Our benchmark evaluates each generated clip from four complementary dimensions: visual quality, action control, state alignment, and mechanics fidelity. All metrics are evaluated on an additional test set of 100 generated samples, evenly covering various mechanics categories.

\begin{itemize}

    \item \textbf{Visual Quality:} Evaluates frame-level and perceptual fidelity by comparing generated videos with reference game-engine outputs:
    \begin{itemize}
        \item[$-$] \textit{SSIM} \cite{wang2004image}: Measures structural similarity after aligning frame size and crop, capturing scene layout, character shapes, and local image structure.
        \item[$-$] \textit{LPIPS} \cite{zhang2018unreasonable}: Measures perceptual similarity using deep visual features. 
    \end{itemize}
    
    \item \textbf{Action Control:} Evaluates how well the generated gameplay follows the commanded player actions. It measures controllability at both the movement and combat levels:
    \begin{itemize}
        \item[$-$] \textit{Movement Accuracy (Move-Acc):} Quantifies alignment between the generated character movement and the input movement direction. We track the character using SAM2.1 \cite{ravi2025sam} and Grounding DINO \cite{liu2024grounding}, and count the movement as successful when the observed displacement satisfies predefined thresholds in a normalized $[0,1]$ coordinate space.
        \item[$-$] \textit{Attack Accuracy (Att-Acc):} Measures the consistency between the generated attack behavior and the intended attack command. We evaluate this using ClipAttackNet, a custom 6-way frame-level attack classifier based on ResNet-18 \cite{he2016deep} and a 4-layer dilated TCN \cite{bai2018empirical}, trained on about 5k clips with a confidence threshold of 0.7.
    \end{itemize}

    \item \textbf{State Alignment:} Measures the accuracy of the generated game states, including the timer, player HP, opponent HP, and both skill meters. For each state variable, we compute the normalized prediction error against the target state trace and report the average distance. We further define the state alignment score as $1-\text{distance}$.

    \item \textbf{Mechanics Fidelity:} Assesses the correctness of mechanics in generated clips. We use Gemini-3.1-Pro \cite{google2026gemini31pro} and GPT-5.5 \cite{openai2026gpt55} as visual judges with a strict prompt and ground-truth reference images, comparing identified mechanics against the target label. This includes, but is not limited to, verifying the correct winner and valid super art execution. Gemini-3.1-Pro processes each video directly, whereas GPT-5.5 evaluates 24 uniformly sampled frames due to limited video input support. Results are reported as mean $\pm$ standard deviation over three runs for more reliable evaluation. Please refer to Appendix \ref{sec:eval} for details.

\end{itemize}

\subsection{Implementation Details}
We adopt Wan2.2-TI2V-5B \cite{wan2025wan} as the base video diffusion model, together with the Wan2.2 VAE and the UMT5-XXL text encoder \cite{chung2023unimax}. The model is trained on our collected dataset of $10,000$ 5-second video clips, where each clip is annotated with player actions, game states, and textual descriptions. Our model is trained for $40,000$ steps using a batch size of $4$ and a learning rate of $5\times10^{-5}$. All video frames are resized to $480 \times 832$. The hyperparameters $N$, $L_S$, $L_V$, $f$, $C$, $K$, $\lambda_{\mathrm{state}}$, and $\lambda_{\mathrm{video}}$ are set to $30$, $25$, $9750$, $25$, $3072$, $11$, $1$, and $1$, respectively.
\begin{table*}[t]
  \centering
  \resizebox{\textwidth}{!}{
  \begin{tabular}{l|ccccccc}
    \toprule
    \raisebox{-0.6\normalbaselineskip}{\textbf{Method}} &
    \multicolumn{2}{c}{\textbf{Visual Quality}} &
    \multicolumn{2}{c}{\textbf{Action Control (\%)}} &
    \raisebox{-0.6\normalbaselineskip}{\textbf{State Alignment} $\uparrow$} &
    \multicolumn{2}{c}{\textbf{Mechanics Fidelity (\%)} $(n=3)$} \\
    \cmidrule(lr){2-3} \cmidrule(lr){4-5} \cmidrule(lr){7-8} 
    &
    SSIM $\uparrow$ & LPIPS $\downarrow$ &
    Move-Acc $\uparrow$ & Att-Acc $\uparrow$ &
    & Gemini-3.1-Pro $\uparrow$ & GPT-5.5 $\uparrow$\\
    \midrule

    \multicolumn{8}{c}{\textbf{Zero-shot Evaluation}} \\
    \midrule
    Matrix-Game 3.0 \cite{wang2026matrix}
    & $0.142$ & $0.673$ & -- & -- & --
    & $42.3_{\pm 0.9}$ & $42.3_{\pm 0.5}$ \\

    LingBot-World \cite{team2026advancing}
    & $0.183$ & $0.601$ & -- & -- & --
    & $44.3_{\pm 0.5}$ & $53.0_{\pm 0.8}$ \\

    LingBot-World 2.0 \cite{gao2026infiniteworldsversatileinteractions}
    & $0.191$ & $0.641$ & -- & -- & --
    & $41.3_{\pm 0.9}$ & $43.0_{\pm 0.8}$ \\

    HY-World 1.5 \cite{sun2025worldplay}
    & $0.172$ & $0.537$ & -- & -- & --
    & $41.0_{\pm 0.8}$ & $43.0_{\pm 1.4}$ \\

    ReactiveGWM \cite{wang2026reactivegwm}
    & $0.340$ & $0.457$ & -- & -- & --
    & $48.0_{\pm 0.7}$ & $43.3_{\pm 1.6}$ \\

    \midrule
    \multicolumn{8}{c}{\textbf{State-aware Fine-tuning}} \\
    \midrule
    Matrix-Game 3.0 \cite{wang2026matrix}
    & $0.240$ & $0.478$ & $40.0$ & $6.67$ & --
    & $48.7_{\pm 1.7}$ & $58.3_{\pm 0.9}$ \\

    HY-World 1.5 \cite{sun2025worldplay}
    & $0.352$ & $0.521$ & $40.0$ & $11.7$ & --
    & $41.7_{\pm 0.5}$ & $38.3_{\pm 0.5}$ \\

    ReactiveGWM \cite{wang2026reactivegwm}
    & $0.376$ & $0.439$ & $\mathbf{95.0}$ & $\mathbf{100.0}$ & --
    & $63.7_{\pm 2.1}$ & $59.7_{\pm 0.9}$ \\

    \rowcolor{cyan!6}
    \textbf{StatePlay}
    & $\mathbf{0.378}$ & $\mathbf{0.424}$
    & $92.5$ & $95.0$
    & $\mathbf{0.947}$
    & $\mathbf{82.3_{\pm 0.5}}$
    & $\mathbf{78.3_{\pm 0.5}}$ \\
    \bottomrule
  \end{tabular}
  }
    \caption{Quantitative comparison of game world models under zero-shot evaluation and state-aware fine-tuning. \textbf{Bold} values indicate the best performance, while ``--'' denotes metrics that are not applicable to a given method.}
  \label{tab:main_results}
\end{table*}

\begin{table*}[t]
  \centering
  \resizebox{\textwidth}{!}{
  \begin{tabular}{cccc|ccccccc}
    \toprule
    \multicolumn{2}{c}{\textbf{Coupling Style}} &
    \multicolumn{2}{c|}{\textbf{State Loss}} &
    \multicolumn{2}{c}{\textbf{Visual Quality}} &
    \multicolumn{2}{c}{\textbf{Action Control (\%)}} &
    \raisebox{-0.6\normalbaselineskip}{\textbf{State Alignment} $\uparrow$} &
    \multicolumn{2}{c}{\textbf{Mechanics Fidelity (\%)} $(n=3)$} \\
    \cmidrule(lr){1-2}
    \cmidrule(lr){3-4}
    \cmidrule(lr){5-6}
    \cmidrule(lr){7-8}
    \cmidrule(lr){10-11}
    Shared & MoT &
    FM & Regression &
    SSIM $\uparrow$ & LPIPS $\downarrow$ &
    Move-Acc $\uparrow$ & Att-Acc $\uparrow$ &
    & Gemini-3.1-Pro $\uparrow$ & GPT-5.5 $\uparrow$ \\
    \midrule

    \cmark &        & \cmark &        
    & $0.309$ & $0.541$
    & $75.0$ & $90.0$
    & $0.804$
    & $52.0_{\pm 0.9}$ & $49.7_{\pm 1.2}$ \\

    \cmark &        &        & \cmark
    & $0.355$ & $0.450$
    & $\mathbf{97.5}$ & $\mathbf{100.0}$
    & $0.943$
    & $64.7_{\pm 2.1}$ & $67.7_{\pm 0.5}$ \\

           & \cmark & \cmark &
    & $0.363$ & $0.430$
    & $85.0$ & $71.7$
    & $0.845$
    & $60.7_{\pm 1.2}$ & $62.0_{\pm 1.4}$ \\

    \rowcolor{cyan!6}
           & \cmark &        & \cmark
    & $\mathbf{0.378}$ & $\mathbf{0.424}$
    & $92.5$ & $95.0$
    & $\mathbf{0.947}$
    & $\mathbf{82.3_{\pm 0.5}}$
    & $\mathbf{78.3_{\pm 0.5}}$ \\

    \bottomrule
  \end{tabular}
  }
    \caption{Comparison of different coupling styles and state loss designs for state-aware world modeling.}
  \label{tab:ablation}
\end{table*}

\subsection{Comparisons with Game World Models}
\subsubsection{Zero-shot Evaluation}
StatePlay is compared with several state-of-the-art game world models in the zero-shot setting first. To ensure a fair comparison, all baseline models are evaluated using their original released architectures and pretrained weights without modification. Since existing game world models neither support our state representation nor share the same action space, we report only visual quality and mechanics fidelity.

As shown in Tab. \ref{tab:main_results}, Matrix-Game 3.0 \cite{wang2026matrix}, LingBot-World \cite{team2026advancing}, LingBot-World 2.0 \cite{gao2026infiniteworldsversatileinteractions} and HY-World 1.5 \cite{sun2025worldplay} produce relatively poor visual quality. Although ReactiveGWM \cite{wang2026reactivegwm}, trained on an in-domain dataset, achieves substantially better visual quality than the other zero-shot baselines, its mechanics fidelity remains below 50\% for both Gemini and GPT evaluation. This result indicates that current game world models can generate visually plausible gameplay but still struggle to faithfully preserve the game rules.

\subsubsection{State-aware Fine-tuning}
We further fine-tune the baseline models on our state-aware dataset to assess whether mechanics-consistent gameplay can emerge from visual supervision alone. To the best of our knowledge, existing open-source game world models neither explicitly model nor predict internal game states; therefore, the state alignment metric is not applicable and is omitted. We also omit fine-tuning results for the LingBot-World model family because they contain over 24B parameters and require substantially more training memory than our 5.75B StatePlay model.

As shown in Tab. \ref{tab:main_results}, StatePlay achieves a state alignment score of 0.947, demonstrating accurate prediction of internal game states. More importantly, it attains the highest mechanics fidelity, achieving 82.3\% under Gemini evaluation and 78.3\% under GPT evaluation, consistently outperforming all baselines over three runs across both evaluators. These results indicate that jointly learning state prediction and frame generation effectively guides the model toward generating gameplay that better follows the underlying game mechanics. Furthermore, introducing explicit state modeling does not compromise the core capabilities of game world models. While significantly improving mechanics fidelity, StatePlay maintains comparable or better visual quality and action control than existing baselines. 

Overall, comparisons in both zero-shot and fine-tuned settings show that existing game world models can generate visually plausible and controllable gameplay but struggle to preserve state-dependent rules when relying on visual modeling alone. StatePlay addresses this limitation by explicitly modeling internal game states and coupling state prediction with frame generation.

\subsection{Ablation Study}
\begin{figure*}[t!]
\centering
    \includegraphics[width=0.95\textwidth]{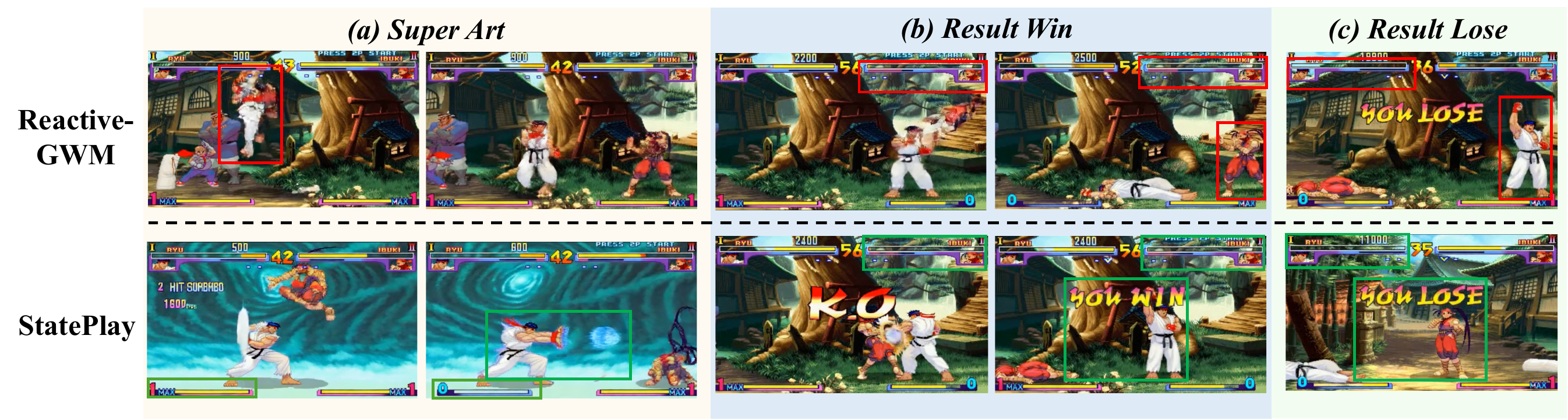}
    \caption{
  Qualitative comparison of three example rollouts (a)--(c) generated by the best-performing baseline, ReactiveGWM, and \textbf{StatePlay}. \textcolor{red}{Red} bounding boxes mark rule-violating rollouts, whereas \textcolor{green!50!black}{green} bounding boxes mark mechanics-consistent rollouts produced by our method. Please zoom in for better visibility.
    }
    \label{fig:visual}
\end{figure*}

\subsubsection{Coupling Style}
\label{subsubsec:couple}
To evaluate the effectiveness of the MoT-style backbone for state-aware world modeling, we compare it with another common architectural choice in world action models: the shared-style backbone \cite{hou2026world}. In the shared-style design, frame tokens and state tokens are processed by a unified backbone, so visual prediction and state prediction are jointly modeled in a common latent space. In contrast, the MoT-style design keeps video and state modeling partially specialized through separate expert branches, while still enabling cross-modal interaction through shared joint attention. From Tab. \ref{tab:ablation}, the performance gap is most pronounced in mechanics fidelity, where the MoT-style backbone outperforms the shared-style backbone by 17.6\%.

This improvement suggests that modality-specific structure is important for state-aware world modeling. Although visual dynamics and game states are tightly coupled, they correspond to intrinsically different representations: video prediction focuses on high-dimensional visual details, while state prediction follows lower-dimensional but rule-sensitive temporal dynamics. Forcing both modalities into a fully shared latent space may interfere with optimization and weaken mechanics modeling. By contrast, the MoT-style backbone allows each modality to preserve specialized representations while maintaining sufficient interaction between visual generation and state prediction.

\subsubsection{State Loss}
\label{subsubsec:state_loss}
Flow matching is commonly adopted in world action models to jointly train frame generation and action prediction. However, we observe that game states have different properties from actions. Actions are usually continuous control signals and may vary freely according to the policy, whereas game states are compact, rule-governed variables with structured temporal patterns. For example, health points typically decrease after attacks, while skill meters usually increase gradually and reset after special moves.

These differences make flow matching less suitable for state prediction. Although effective for high-dimensional visual generation, flow matching introduces random perturbations to low-dimensional state variables and requires the model to learn a transport vector field over their distributions. This formulation may unnecessarily complicate state prediction, whose transitions are largely determined by explicit mechanics and can be modeled more directly through temporal regression. Our experimental results from Tab. \ref{tab:ablation} further show that supervising states with a direct regression loss consistently outperforms flow matching, yielding improvements of 12.1\% in state alignment and 21.6\% in mechanics fidelity. Additional ablation studies are provided in Appendix \ref{sec:more_abla}.


\subsection{Qualitative Visualization}
We compare the qualitative results of the best-performing baseline, ReactiveGWM, and our StatePlay. For a fair comparison, both methods are given the same action sequence, text prompt, initial frame, and initial state as input.

As shown in Fig.~\ref{fig:visual}(a), StatePlay successfully performs the super art when the skill meter is full and correctly resets the meter to 0 after execution. In contrast, ReactiveGWM fails to trigger the super art even when the skill meter reaches the required threshold. It also produces blurred visual artifacts, where the two characters become visually entangled.

Fig.~\ref{fig:visual}(b) shows a game-ending case where the NPC's health point reaches zero. ReactiveGWM produces an invalid rollout in which the NPC remains standing and attacking despite having zero health, whereas StatePlay correctly generates the ``You Win'' result. Fig.~\ref{fig:visual}(c) shows the opposite case, where the player's health point reaches zero. ReactiveGWM incorrectly depicts the player as winning, with the player celebrating and the NPC lying on the ground, while StatePlay generates the correct losing outcome consistent with the game rule.

These visualization results further show that game world models without explicit state modeling, such as ReactiveGWM, tend to overfit to pixel-level visual dynamics and fail to capture valid game mechanics. In contrast, our state-aware design enables StatePlay to predict internal states more accurately and generate mechanics-consistent rollouts. Please refer to the Appendix \ref{sec:more_visual} for more visualization results.

\section{Conclusions and Discussions}
In this paper, we introduce \textbf{StatePlay}, a state-aware game world model that incorporates explicit internal state prediction into game frame generation. Unlike previous game world models that mainly focus on visual dynamics conditioned on player actions, StatePlay models states such as timers, health points, and skill meters as part of the generative process. By coupling state prediction with frame generation, our method produces rollouts that are both visually plausible and more consistent with the underlying game mechanics. Experiments on an intuitive fighting game demonstrate that StatePlay accurately predicts internal states and improves mechanics fidelity while preserving action controllability and visual quality. These results establish the importance of state-aware modeling and provide an initial viable framework for more complete and playable game world models.

However, as publicly available datasets with synchronized state, frame and action remain limited, we focus our initial evaluation on \textit{Street Fighter 3}. Broader validation across game genres requires larger and more diverse state-aware datasets. Given the benefits of state modeling for mechanics-consistent generation, we hope this work encourages the development of datasets with explicit state annotations across a wider range of games, enabling state-aware game world models to scale and generalize across diverse rule systems.


\clearpage
\setcounter{page}{1}
\setcounter{section}{0} 
\renewcommand{\thesection}{\Alph{section}}

\section{Experimental Model Information}
\label{sec:model_info}
Tab. \ref{tab:model_configurations} summarizes the model configuration used in our experiments. StatePlay adds a lightweight 0.76B-parameter state branch while consistently outperforming the game world model baselines. Since most models are built on base diffusion models, text encoders, and VAEs from the same model family, these results further demonstrate the effectiveness of our MoT-style architecture for state-aware game modeling.
\begin{table}[h]
    \centering
    \scriptsize
    \setlength{\tabcolsep}{2.5pt}
    \renewcommand{\arraystretch}{1.12}
    \begin{tabularx}{\columnwidth}{
        >{\raggedright\arraybackslash}p{1.55cm}
        >{\centering\arraybackslash}p{0.85cm}
        >{\raggedright\arraybackslash}X
        >{\raggedright\arraybackslash}p{1.45cm}
        >{\raggedright\arraybackslash}p{1.25cm}
    }
        \toprule
        \textbf{Model} &
        \textbf{Params.} &
        \textbf{Base Model} &
        \textbf{Text Encoder} &
        \textbf{VAE} \\
        \midrule

        Matrix-Game 3.0 &
        6.312B &
        Wan2.2-TI2V-5B &
        UMT5-XXL &
        Wan2.2 \\
        \cmidrule(lr){1-5}

        LingBot-World &
        37.089B &
        Wan2.1-I2V-A14B with high-/low-noise DiTs &
        UMT5-XXL &
        Wan2.1 \\
        \cmidrule(lr){1-5}

        LingBot-World 2.0 &
        24.352B &
        Wan2.1-I2V-A14B &
        UMT5-XXL &
        Wan2.1 \\
        \cmidrule(lr){1-5}

        HY-World 1.5 &
        18.11B &
        HunyuanVideo-1.5 480P-I2V &
        Qwen2.5-VL-7B-Instruct &
        HYVideo-1.5 \\
        \cmidrule(lr){1-5}
        
        ReactiveGWM &
        5.001B &
        Wan2.2-TI2V-5B &
        UMT5-XXL &
        Wan2.2 \\
        \cmidrule(lr){1-5}

        \textbf{StatePlay} &
        \textbf{5.759B} &
        Wan2.2-TI2V-5B &
        UMT5-XXL &
        Wan2.2 \\
        \bottomrule
    \end{tabularx}
    \caption{Architecture configurations of the evaluated game world models.}
    \label{tab:model_configurations}
\end{table}

\section{State-aware Dataset Construction}
\label{sec:dataset}
\subsection{Mechanics Distribution Balancing Criteria}
\label{subsec:mechanics_distribution}
To ensure balanced coverage of diverse mechanics scenarios, we construct the training dataset according to the distribution shown in Tab. \ref{tab:data_categories}. A clip is assigned to a specific category only when it satisfies both the corresponding visual and state conditions, ensuring consistency between the observed gameplay and the underlying mechanics. We collect 1,000 clips for each of the four state-critical categories and 6,000 clips for normal gameplay, resulting in 10,000 training samples.
\begin{table}[h]
    \centering
    \scriptsize
    \setlength{\tabcolsep}{2.5pt}
    \renewcommand{\arraystretch}{1.08}
    \begin{tabularx}{\columnwidth}{
        >{\raggedright\arraybackslash}p{1.25cm}
        >{\centering\arraybackslash}p{0.60cm}
        >{\raggedright\arraybackslash}X
        >{\raggedright\arraybackslash}X
    }
        \toprule
        \textbf{Category} &
        \textbf{\#} &
        \textbf{Visual condition} &
        \textbf{State condition} \\
        \midrule

        \textit{Result Win} &
        1k &
        ``You Win'' in final frame &
        Player HP $>0$, NPC HP $=0$ \\
        \cmidrule(lr){1-4}

        \textit{Result Lose} &
        1k &
        ``You Lose'' in final frame &
        Player HP $=0$, NPC HP $>0$ \\
        \cmidrule(lr){1-4}

        \textit{Macro Success} &
        1k &
        Super art is executed &
        Macro input; Player skill meter $\geq \tau$ \\
        \cmidrule(lr){1-4}

        \textit{Macro Fail} &
        1k &
        No super art is executed &
        Macro input; Player skill meter $<\tau$ \\
        \cmidrule(lr){1-4}

        \textit{Normal} &
        6k &
        No terminal or super art event &
        None \\
        \midrule

        \textbf{Total} &
        \textbf{10k} &
        \multicolumn{2}{l}{40\% state-critical, 60\% normal} \\
        \bottomrule
    \end{tabularx}
    \caption{Dataset categories and classification criteria.}
    \label{tab:data_categories}
\end{table}

\subsection{NPC Strategy Annotation}

\label{subsec:NPC_anno}

To encourage NPCs to engage players strategically in generated rollouts, we follow ReactiveGWM \cite{wang2026reactivegwm} and use Gemini-3.1-Pro \cite{google2026gemini31pro} to classify NPC behavior into three strategies: \textit{offense}, \textit{control}, and \textit{defense}. The detailed annotation prompt is shown in Fig. \ref{fig:npc_prompt}. The resulting strategy label is included in the training dataset alongside the corresponding video clip, player actions, and game states.

\begin{figure}[h]
\centering

\begin{tcolorbox}[
    width=\columnwidth,
    colback=white,
    colframe=gray!35,
    colbacktitle=gray!25,
    coltitle=black,
    title=\textbf{NPC Strategy Prompt},
    fonttitle=\normalsize\bfseries,
    boxrule=0.8pt,
    arc=2mm,
    left=2mm,
    right=2mm,
    top=1.5mm,
    bottom=1.5mm
]

You are given a 5-second gameplay clip from a fighting game. Analyze only the NPC's visible behavior and classify its strategy as \textit{offense}, \textit{control}, or \textit{defense}. Do not infer hidden intentions or use information that is not visually observable.

First, record the following factual observations: whether the NPC performs punches, kicks, jumping attacks, throws, special melee attacks, or projectile attacks; whether it advances toward the player; whether it takes damage; whether it applies sustained close-range pressure; whether it crouches or guards; the dominant engagement range (\textit{close}, \textit{mid}, or \textit{far}); and which character attacks more frequently.

Then assign exactly one strategy:

\begin{itemize}

    \item \textbf{Offense}: the NPC advances, repeatedly attacks at close range, applies sustained pressure, and initiates more attacks than the player.

    \item \textbf{Control}: the NPC maintains mid or long range, avoids close-range exchanges, and uses one or more projectile attacks to restrict the player's movement.

    \item \textbf{Defense}: the NPC performs few or no attacks, does not advance, and mainly guards, crouches, retreats, blocks, evades, or reacts to damage.
\end{itemize}

\end{tcolorbox}

\caption{NPC strategy annotation prompt for classifying fighting behaviors and incorporating diverse strategy descriptions into the dataset.}
\label{fig:npc_prompt}
\end{figure}

\begin{figure*}[!t]

\centering
\begin{tcolorbox}[
    width=0.96\textwidth,
    colback=white,
    colframe=gray!35,
    colbacktitle=gray!25,
    coltitle=black,
    title=\textbf{Evaluation Prompt (Mechanics Fidelity)},
    fonttitle=\large\bfseries,
    boxrule=0.8pt,
    arc=3mm,
    left=3mm,
    right=3mm,
    top=2mm,
    bottom=2mm
]
You are evaluating a generated Street Fighter 3 gameplay clip.

\textbf{Allowed output categories:} normal, macro\_success, macro\_fail, result\_win and result\_lose

You must classify what is actually visible in the generated video, not what the filename or ground-truth says.

\textbf{Important visual rules:}

\begin{enumerate}[
    leftmargin=6mm,
    label=\arabic*.,
    itemsep=2pt,
    topsep=1pt,
    parsep=0pt
]
    \item \textbf{result\_win:}\\
    Choose result\_win only when the video visibly shows the player's win result screen/text, especially ``YOU WIN'', or an unmistakable win-result screen matching the attached result\_win reference. A K.O. frame, opponent falling, victory pose, depleted opponent HP, or a final hit is NOT enough by itself if ``YOU WIN'' / win result screen is not visible.

    \item \textbf{result\_lose:}\\
    Choose result\_lose only when the video visibly shows the player's lose result screen/text, especially ``YOU LOSE'', or an unmistakable lose-result screen matching the attached result\_lose reference. A K.O. frame, player falling, depleted player HP, or the opponent winning is NOT enough by itself if ``YOU LOSE'' / lose result screen is not visible.

    \item \textbf{macro\_success:}\\
    Choose macro\_success when a character super-art / special macro sequence is visibly executed and connects or clearly affects the opponent. Look for large super-art effects, dramatic freeze/flash, cinematic hit sequence, multiple-hit super move, or the attached macro reference. It does not need to end the round.

    \item \textbf{macro\_fail:}\\
    Choose macro\_fail when a super-art / macro sequence is visibly attempted but misses, is blocked, whiffs, is interrupted, or produces no clear hit/effect on the opponent.

    \item \textbf{normal:}\\
    Choose normal for ordinary gameplay, ambiguous fighting, jump/attack/block movement, throws, regular hits, fireballs, K.O. without visible YOU WIN/YOU LOSE result text, or if none of the above special/result states is clearly visible.
\end{enumerate}

\textbf{Decision priority:}
\begin{itemize}[
    leftmargin=5mm,
    itemsep=0pt,
    topsep=1pt,
    parsep=0pt
]
    \item If ``YOU WIN'' is visible, answer result\_win.
    \item Else if ``YOU LOSE'' is visible, answer result\_lose.
    \item Else if a super-art / macro clearly hits or affects the opponent, answer macro\_success.
    \item Else if a super-art / macro is clearly attempted but fails/misses/gets blocked, answer macro\_fail.
    \item Else answer normal.
\end{itemize}

\textbf{The attached reference images are, in order:} result\_win reference, result\_lose reference, macro/super-art reference
\textbf{Return one JSON object only.} Do not return a list. Do not include markdown.
{\scriptsize
\begin{verbatim}
{
  "predicted_state": "normal|macro_success|macro_fail|result_win|result_lose",
  "matches_gt": true,
  "confidence": 0.0,
  "evidence": {
    "you_win_visible": true,
    "you_lose_visible": false,
    "super_art_visible": false,
    "super_art_hits": false
  },
  "reason": "one short sentence naming the decisive visible evidence"
}
\end{verbatim}
}

\end{tcolorbox}

 \begin{subfigure}[t]{0.325\textwidth}
        \centering
        \includegraphics[width=\linewidth,height=0.55\textwidth,
                         keepaspectratio]{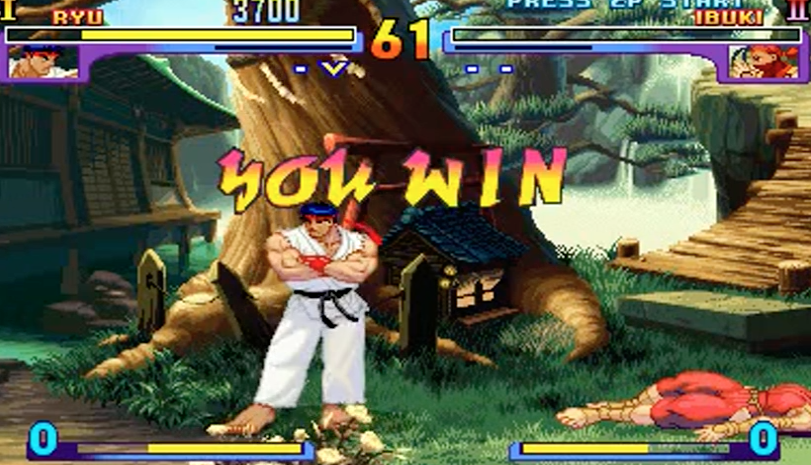}
        \label{fig:ref-win}
    \end{subfigure}
    \hfill
    \begin{subfigure}[t]{0.325\textwidth}
        \centering
        \includegraphics[width=\linewidth,height=0.55\textwidth,
                         keepaspectratio]{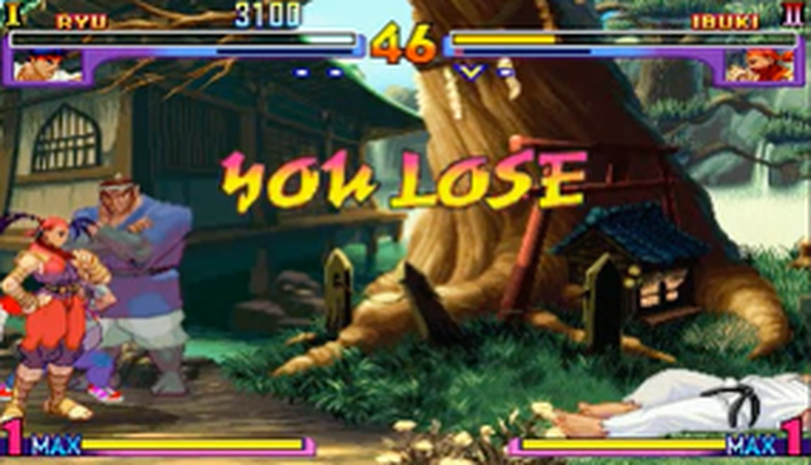}
        \label{fig:ref-lose}
    \end{subfigure}
    \hfill
    \begin{subfigure}[t]{0.325\textwidth}
        \centering
        \includegraphics[width=\linewidth,height=0.55\textwidth,
                         keepaspectratio]{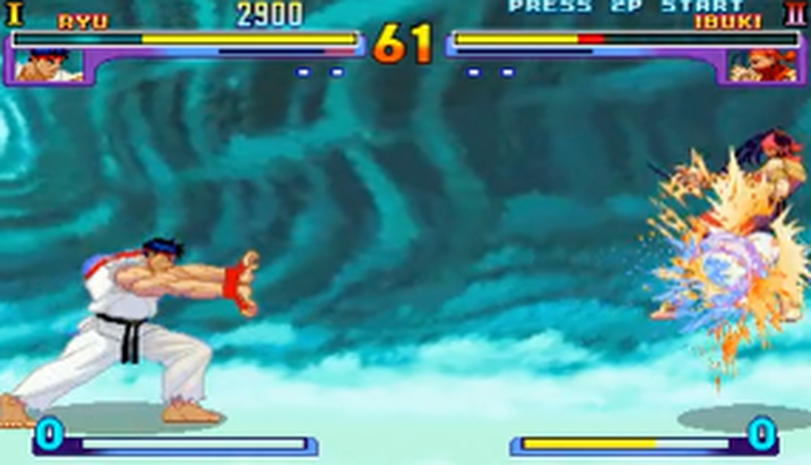}
        \label{fig:ref-macro}
    \end{subfigure}
    \caption{Evaluation prompt for mechanics fidelity (top), with three ground-truth reference images (bottom).}
    \label{fig:eval_prompt}

\end{figure*}

\begin{figure*}[!t]
\centering
    \includegraphics[width=0.95\textwidth]{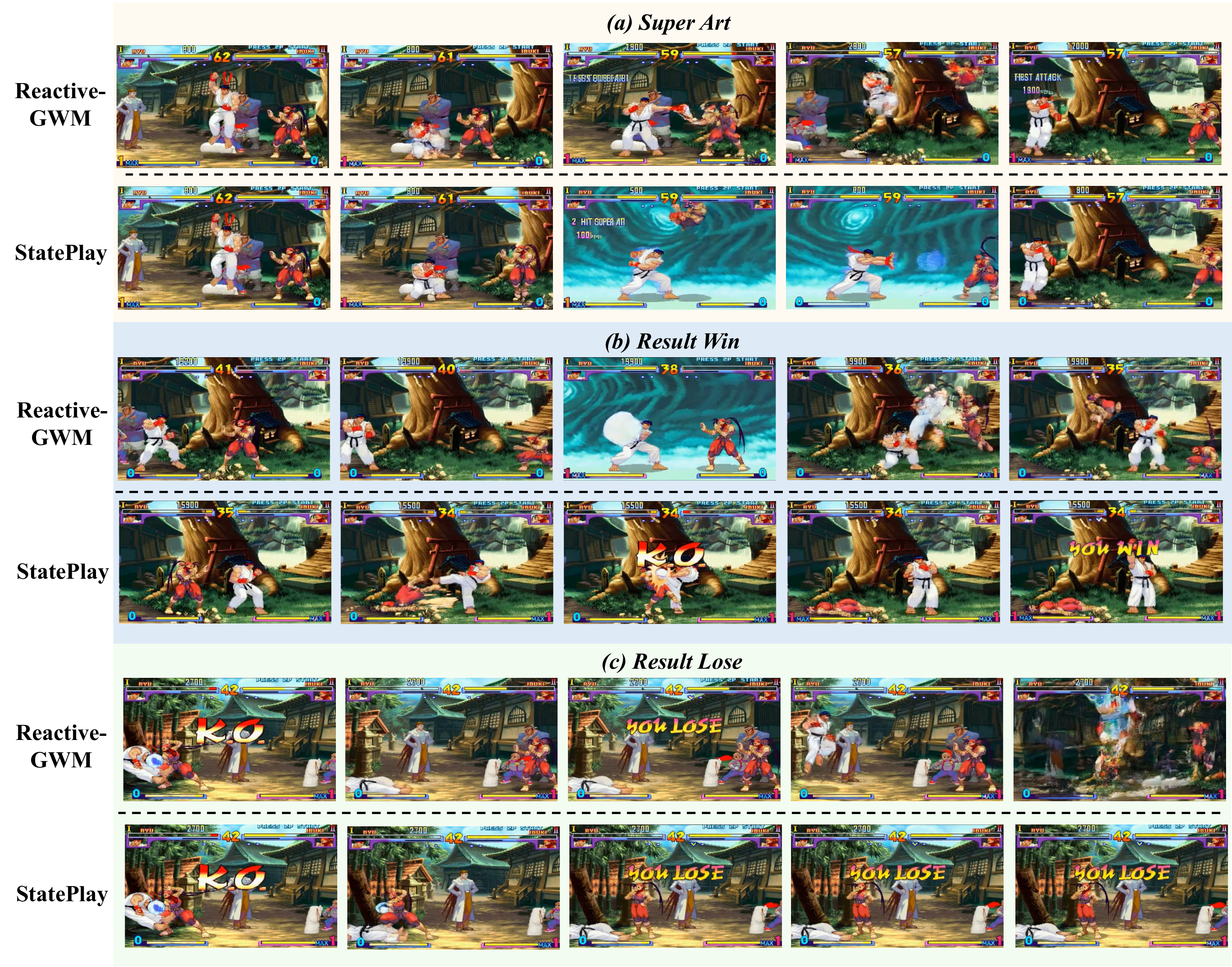}
    \caption{
  Qualitative comparison of three example rollouts (a)--(c) generated by the best-performing baseline, ReactiveGWM, and \textbf{StatePlay}. Please zoom in for better visibility.
    }
    \label{fig:more_visual}
\end{figure*}

\section{Failure Cases}
Although StatePlay accurately predicts the internal states required for valid mechanics, pixel-level indicators such as health bars and skill meters may still show occasional inconsistencies. Visual quality can also degrade when multiple mechanics occur simultaneously, such as a super art coinciding with the end of a match. Stronger visual generation capabilities may help address these limitations.

\section{Details of Mechanics Fidelity Evaluation}
\label{sec:eval}
To robustly evaluate mechanics fidelity, we prompt Gemini-3.1-Pro and GPT-5.5 with explicit visual rules, a fixed decision priority, and reference images to determine whether each generated rollout follows the ground-truth mechanics. Win and loss are recognized only when the corresponding result screen is visible, while macro success requires a clearly executed super art that successfully hits the opponent. The complete prompt and reference images are shown in Fig.~\ref{fig:eval_prompt}.

\section{State Input Comparisons}
\label{sec:more_abla}
Since StatePlay uses a regression objective for state prediction, we compare two ways of constructing the state input during training: broadcasting the initial clean state across all timesteps and adding noise to the state sequence. 

As shown in Tab. \ref{tab:state_input_ablation}, the noise-based input consistently outperforms the broadcasting baseline in both state alignment and mechanics fidelity. We hypothesize that broadcasting the same clean state across all timesteps produces an overly static initialization, which hinders the learning of temporally evolving frame dynamics. Therefore, following prior world action models \cite{hu2026bagelvla,ye2026world}, we perturb both the video latents and raw states during training.

\begin{table}[h]
    \centering
    \scriptsize
    \setlength{\tabcolsep}{7pt}
    \renewcommand{\arraystretch}{1.12}
    \begin{tabular}{cc|ccc}
        \toprule
        \multicolumn{2}{c|}{\textbf{State Input}} &
        \raisebox{-0.6\normalbaselineskip}{\textbf{State Alignment} $\uparrow$} &
        \multicolumn{2}{c}{\textbf{Mechanics Fidelity (\%)}} \\
        \cmidrule(lr){1-2}
        \cmidrule(lr){4-5}
        Broadcast & Noise & &
        Gemini $\uparrow$ & GPT $\uparrow$ \\
        \midrule

        \cmark &        
        & $0.938$
        & $78.3_{\pm 2.4}$
        & $76.0_{\pm 0.8}$ \\

        \rowcolor{cyan!6}
        & \cmark
        & $\mathbf{0.947}$
        & $\mathbf{82.3_{\pm 0.5}}$
        & $\mathbf{78.3_{\pm 0.5}}$ \\

        \bottomrule
    \end{tabular}
    \caption{Comparison of different state input designs.}
    \label{tab:state_input_ablation}
\end{table}
\section{More Qualitative Comparisons}
\label{sec:more_visual}
We provide additional qualitative comparisons in Fig. \ref{fig:more_visual}. ReactiveGWM and StatePlay are fine-tuned on the same state-aware dataset for $40{,}000$ steps. Despite this identical training setup, ReactiveGWM, which models only visual observations without explicitly predicting game states, frequently generates rollouts that violate the underlying game mechanics. Specifically, Fig. \ref{fig:more_visual}(a) shows that it fails to execute a super art even when the skill meter is full. In Fig. \ref{fig:more_visual}(b), it does not display the game result after the NPC's health reaches zero. In Fig. \ref{fig:more_visual}(c), although the ``You Lose'' message appears, the subsequent frames become blurred and distorted instead of transitioning to a proper game-ending screen. In contrast, StatePlay generates mechanics-consistent rollouts by explicitly modeling the underlying game states.


\end{document}